\documentclass[fleqn,10pt]{wlscirep}
\usepackage{amsmath}
\usepackage{amssymb}
\usepackage[utf8]{inputenc}
\usepackage[T1]{fontenc}
\usepackage{hyperref}
\usepackage{float}

\newcommand{\DesignI}{\textbf{\texttt{NeuroShield }}}
\newcommand{\DesignII}{\textbf{\texttt{NeuroShield}}}

\title{NeuroShield: A Neuro-Symbolic Framework for Adversarial Robustness}

\author[1]{Ali Shafiee Sarvestani}
\author[2]{Jason Schmidt}
\author[1*]{Arman Roohi}
\affil[1]{University of Illinois Chicago, Department of Electrical and Computer Engineering, Chicago, IL, 60607, USA}
\affil[2]{University of Illinois Chicago, Department of Computer Science, Chicago, IL, 60607, USA}

\affil[*]{aroohi@uic.edu}

\begin{abstract}
Adversarial vulnerability and lack of interpretability are critical limitations of deep neural networks, especially in safety-sensitive settings such as autonomous driving. We introduce \DesignII, a neuro-symbolic framework that integrates symbolic rule supervision into neural networks to enhance both adversarial robustness and explainability. Domain knowledge is encoded as logical constraints over appearance attributes such as shape and color, and enforced through semantic and symbolic logic losses applied during training. Using the GTSRB dataset, we evaluate robustness against FGSM and PGD attacks at a standard $\ell_\infty$ perturbation budget of $\varepsilon = 8/255$. Relative to clean training, standard adversarial training provides modest improvements in robustness ($\sim$10 percentage points). Conversely, our FGSM-Neuro-Symbolic and PGD-Neuro-Symbolic models achieve substantially larger gains, improving adversarial accuracy by 18.1\% and 17.35\% over their corresponding adversarial-training baselines, representing roughly a three-fold larger robustness gain than standard adversarial training provides when both are measured relative to the same clean-training baseline, without reducing clean-sample accuracy. Compared to transformer-based defenses such as LNL-MoEx, which require heavy architectures and extensive data augmentation, our PGD-Neuro-Symbolic variant attains comparable or superior robustness using a ResNet18 backbone trained for 10 epochs. These results show that symbolic reasoning offers an effective path to robust and interpretable AI.

\end{abstract}
\begin{document}

\flushbottom
\maketitle
%
%
\thispagestyle{empty}


\section*{Introduction}

The black-box nature of deep learning models makes them vulnerable to adversarial attacks and failure modes. Their lack of transparency limits human interpretability and supervision. Traditional human oversight is primarily limited to dataset preparation and pre-processing, which is insufficient for safety-sensitive tasks requiring trust and explainability.

\begin{figure}[t]
  \centering
  \includegraphics[width=0.45\textwidth]{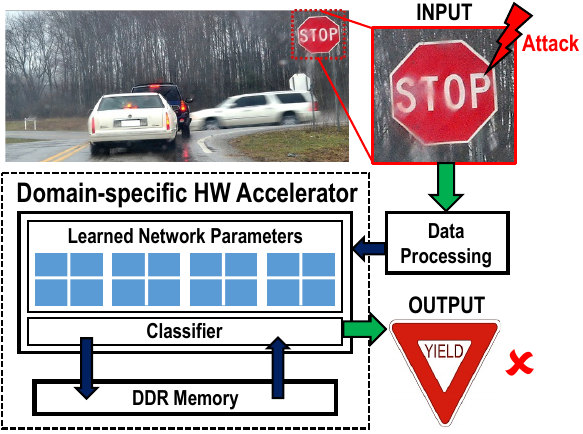}
  \caption{Soft errors in an NN accelerator can lead to image misclassification in a self-driving car, potentially causing it to accelerate instead of braking abruptly. (adopted from \cite{ref28}).}
  \label{fig:soft_error}
\end{figure}

Neuro-Symbolic AI (NSAI) augments deep neural networks (DNNs) with symbolic reasoning to enhance interpretability, trust, and robustness~\cite{ref5}. DNNs can extract abstract features from large datasets and learn complex decision boundaries~\cite{ref1}. However, their inner mechanisms remain opaque, forcing users to trust predictions without meaningful insight into the decision-making process. Such opacity can cause harmful errors in real-world deployments.
NSAI addresses this by incorporating domain-specific symbolic rules into neural models. These symbolic components act as expert supervisors, verifying whether the model's output adheres to known logical relationships. For example, in the classification of traffic signs, a symbolic rule might require that a ``STOP'' sign be red and octagonal. If a model predicts a class without detecting these attributes, the symbolic module can flag the inconsistency, Fig.\ref{fig:soft_error}.
Depending on the integration strategy, the symbolic component may function as a post-hoc verifier, a parallel logic engine, or an embedded structure that influences training dynamics. Some approaches integrate logic into the loss; others use symbolic rules to filter or re-rank outputs~\cite{ref2,taha2002symbolic}. NSAI systems may rely on predefined rules, analogous to supervised learning, or dynamically extract new rules during training through data-driven induction mechanisms.
Despite strong performance, DNNs remain highly susceptible to adversarial threats. These threats arise from small, intentional perturbations, often imperceptible to the human eye, designed to mislead the model. This vulnerability stems from reliance on superficial statistical correlations rather than semantic reasoning. Thus, even well-trained models can be fooled by inputs that look nearly identical to clean samples. Adversarial attacks can occur during both training and inference. Some corrupt training by injecting malicious samples; others target inference by inducing confident but incorrect predictions~\cite{ref3}. These attacks may be designed to be broad and random, or highly targeted toward specific outputs. In either case, they exploit the lack of reasoning and interpretability, revealing the brittleness of neural models.

In response, various defense strategies have been developed to make the models more resilient. These generally involve either adjusting the training process to expose the model to noisy or perturbed input, modifying inputs at test time to suppress adversarial patterns, or altering the model architecture to obscure the gradients that attackers rely on. Some approaches attempt to guarantee robustness through formal verification techniques, while others aim to reduce the model’s sensitivity to input changes~\cite{ref3}. However, all of these approaches operate purely in the neural domain and remain limited by training data.
This limitation motivates integrating symbolic reasoning into learning. Unlike purely statistical models, neuro-symbolic systems can evaluate predictions through the lens of logic and structured knowledge. When encountering an adversarial input, such a system is not solely dependent on learned patterns; it can also rely on logical constraints representing domain knowledge. These constraints help ensure that output remains consistent with fundamental relationships, such as a known traffic sign's shape and color requirements. Even if a perturbed input misleads the neural component, the symbolic module acts as a semantic filter, identifying contradictions. In this way, reasoning capabilities offer an additional layer of defense, one that enhances robustness by establishing predictions in interpretable and verifiable rules rather than purely statistical features.

This paper presents \DesignII, an NSAI framework designed to improve the explainability and robustness of neural networks by integrating symbolic reasoning into the learning pipeline.
The remainder of this paper is organized as follows. Section 1 reviews related work on neuro-symbolic learning and adversarial robustness. Section 2 presents our rule-based supervision framework. Sections 3 and 4 detail the backbone modifications and the training pipeline. Section 6 evaluates the robustness, interpretability, and efficiency of our model.

\section{Related Works}
Since our method integrates symbolic reasoning into neural networks for adversarial robustness, we organize the related work into two parts: symbolic reasoning, which reviews neuro-symbolic frameworks that embed logic into neural learning, and adversarial defenses, which survey techniques for defending against adversarial attacks.

\subsection{Symbolic Reasoning} 
Recent advances in neuro-symbolic AI have explored integrating symbolic reasoning with neural network learning to improve interpretability, consistency, and generalization. Logic Tensor Networks (LTNs)~\cite{ref2} combine first-order logic with deep learning by embedding logical symbols (constants, functions, predicates) into real-valued tensors through a differentiable logic system called Real Logic. LTNs support fuzzy truth values, logical operations, and quantifiers, and have been applied to classification, regression, and semi-supervised tasks. Neural Logic Machines (NLMs)~\cite{ref6} extend this idea with a differentiable architecture for relational reasoning, using weight sharing and pooling to learn logic-like rules and generalize across tasks such as transitive reasoning and program induction.

Neuro-Symbolic Forward Reasoning (NSFR)~\cite{ref7} integrates symbolic logic with perception by extracting object-centric representations from images and mapping them to logical atoms through predicate classifiers. These atoms are processed by a differentiable forward-chaining engine that performs multi-step inference over fuzzy truth values, achieving interpretable results on visual reasoning tasks such as Kandinsky Patterns and CLEVR-Hans. Unlike methods that rely solely on latent features, NSFR emphasizes rule-driven behavior grounded in perception. Semantic loss~\cite{ref8} provides another way to embed symbolic constraints by defining a differentiable loss over propositional logic formulas. It penalizes confidence in logic-violating outputs and has been effective in semi-supervised classification and structured prediction tasks such as path generation and preference ranking.

Finally, neuro-symbolic verification~\cite{ref9} applies symbolic reasoning to model verification by combining neural networks with logic-based assertion checking. A specification network defines semantic conditions, and verification rules are encoded in a formal assertion language and checked using SMT (Satisfiability Modulo Theories) solvers, enabling validation of semantic and multi-network properties for safety-critical systems.

\subsection{Adversarial Robustness} 
In this section, we review defense strategies designed to improve robustness against adversarial attacks, including adversarial training, input transformation, adversarial detection, and certified defenses.
Recent studies show that adversarial training remains one of the most effective defense mechanisms. Fast Gradient Sign Method (FGSM) introduced in~\cite{ref10}, generates adversarial examples via a single gradient-sign step and showed substantial robustness gains on MNIST, recovering over 60 percentage points of adversarial accuracy while maintaining high clean accuracy. However, FGSM training may cause gradient masking and fails against stronger iterative attacks. To overcome these limitations, in~\cite{ref11}, adversarial training was formalized as a robust optimization problem, and Projected Gradient Descent (PGD) was introduced as a universal first-order adversary. PGD-based adversarial training demonstrated significantly improved robustness, achieving up to 89\% precision under the PGD attack on MNIST, and became the de facto baseline for adversarial defenses. Nonetheless, PGD training is computationally intensive and can overfit the specific attack used during training, reducing its effectiveness against unseen or adaptive threats.

Beyond adversarial training, input transformation is a practical defense that attempts to remove perturbations before classification. Common techniques include JPEG compression, bit-depth reduction, pixel dropout, and total variation minimization. These methods are model-agnostic and easy to apply, but often reduce clean accuracy and remain vulnerable to adaptive attacks. Many input-based defenses have been shown to rely on gradient obfuscation, which can give a false sense of security and be circumvented by carefully designed attacks~\cite{ref12}.
Another defense direction is adversarial example detection, which uses auxiliary models or statistical tests to identify adversarial inputs by examining feature activations, gradients, or output distributions. While sometimes effective, detection approaches often fail to generalize across datasets or attacks and are particularly weak in white-box settings where the detector can be directly targeted~\cite{ref13}.
Lastly, certified defenses aim to provide formal robustness guarantees by designing models that are provably resistant to perturbations within a given norm ball. Techniques such as randomized smoothing, Lipschitz regularization, and convex relaxation fall into this category. Although these approaches offer provable robustness, they are computationally expensive and struggle to scale to large models and complex datasets~\cite{ref14}.

\begin{figure*}[t]
  \centering
  \includegraphics[width=0.9\textwidth]{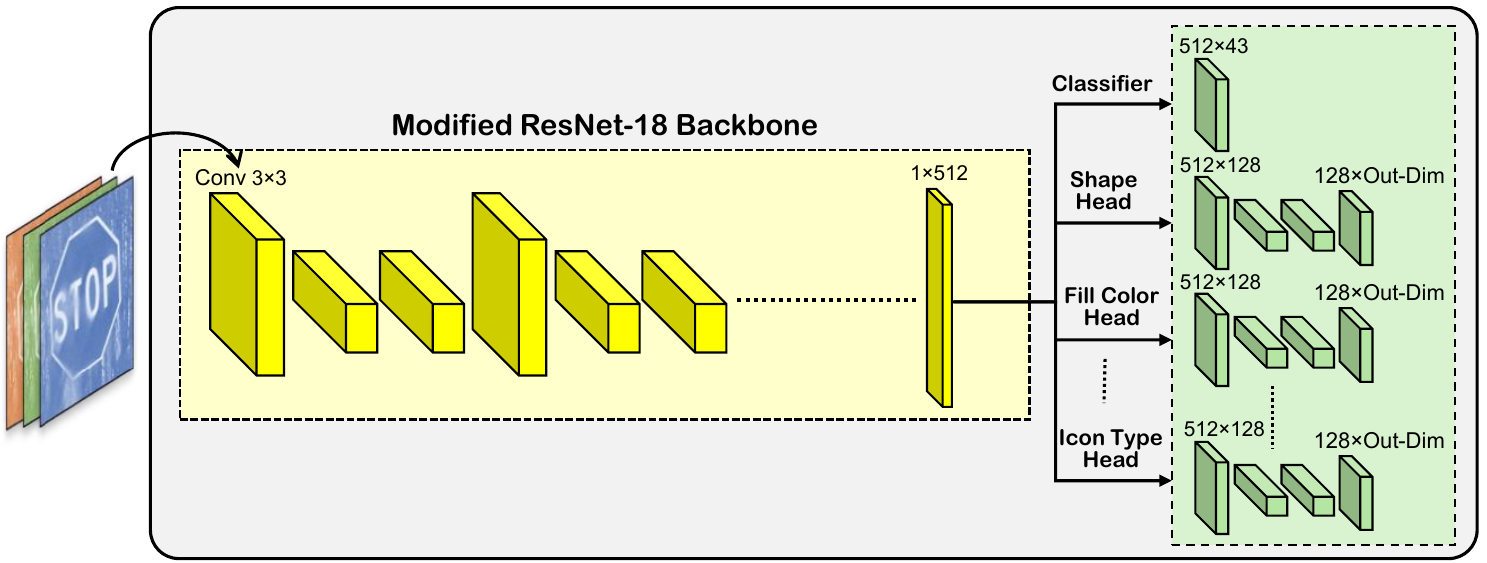}
  \caption{Modified ResNet18 architecture used in our framework. The network preserves high spatial resolution in early layers by replacing the initial 7×7 convolution with a 3×3 kernel and removing the max pooling layer. The final fully connected layer is removed, and the resulting 512-dimensional feature vector is shared between the traffic sign classifier and symbolic attribute heads.}
  \label{fig:resnet-arch}
\end{figure*}

\section{Rule Representation and Symbolic Supervision}
Herein, we represent prior knowledge using symbolic rules that associate each traffic sign class with its expected attributes, including shape, fill and border color, icon or digit type, and category. These rules form a lightweight knowledge base that enables the model to interpret what a prediction semantically entails. For instance, a STOP sign should be red, octagonal, contain white text with a central “STOP” icon, have a white border, and belong to the priority category. These associations are defined statically using class-wise dictionaries (e.g., \texttt{class\_shape\_map}, \texttt{class\_color\_parts}) that encode expected values for each symbolic property.
Each rule is translated into a soft implication of the form:
\begin{equation}
\text{if class = } c \Rightarrow \text{shape = } s,\ \text{color(fill) = } f,\ \text{category = } k,\ \ldots
\tag{1}
\end{equation}
We incorporate two forms of symbolic regularization to improve both interpretability and robustness. First, inspired by semantic loss~\cite{ref8}, we define a simplified variant that encourages the predicted class distribution to align with a group of symbolically equivalent classes. Instead of forcing the model to always predict one exact label, this loss allows it to distribute confidence across all symbolically valid classes (e.g., all speed limit signs). This makes the model's mistakes more interpretable.
Given the ground-truth label \( y \), we identify all classes that share the same symbolic attributes, such as shape, color, icon type, and category, and construct a soft target distribution \( q \):
\begin{equation}
q_i = \begin{cases}
\frac{1}{|\mathcal{S}(y)|} & \text{if } i \in \mathcal{S}(y) \\
0 & \text{otherwise}
\end{cases}
\tag{2}
\end{equation}
Here, \( \mathcal{S}(y) \) denotes the set of symbolically equivalent classes for the label \( y \), and \( q \in \mathbb{R}^{43} \) defines the soft target distribution over classes.

The semantic loss is computed as the Kullback--Leibler divergence between this symbolic target \( q \) and the predicted class distribution \( p_{\text{cls}} \):
\begin{equation}
\mathcal{L}_{\text{semantic}} = \text{KL}(q \parallel p_{\text{cls}})
\tag{3}
\end{equation}
Kullback-Leibler (KL) divergence is a measure of how one probability distribution diverges from a second expected distribution. In this context, it penalizes the model when it assigns high confidence to classes that are not symbolically consistent with the true label.

Second, we introduce a symbolic logic loss that enforces consistency between the predicted class and its associated symbolic components. Rather than enforcing hard restrictions, we use soft implications inspired by the Pylon framework~\cite{ref15}. For each component, we compute a soft truth value:
\begin{equation}
T(A \Rightarrow B) = \begin{cases}
\frac{p_B}{p_A + \varepsilon} & \text{if } p_A > p_B \\
1 & \text{otherwise}
\end{cases}
\tag{4}
\end{equation}
Here, \( p_A \) is the confidence in the predicted class, and \( p_B \) is the confidence in the associated symbolic attribute. \( T(A \Rightarrow B) \in [0, 1] \) represents the degree to which the symbolic implication is satisfied. This formulation ensures that when the model confidently predicts a class (i.e., \( p_A \) is high), it is also held responsible for assigning high confidence to the corresponding symbolic attributes (i.e., \( p_B \)). If \( p_B \) is significantly lower than \( p_A \), the implication truth drops, signaling a violation and resulting in a higher loss. In contrast, if \( p_A \) is low or \( p_B \geq p_A \), the implication is considered satisfied and no penalty is applied. This soft implication mechanism, as defined in Equation~(4), enforces the principle that confident predictions must be semantically coherent. The resulting symbolic logic loss penalizes these violations using a logarithmic function.
\begin{equation}
\mathcal{L}_{\text{logic}} = -\log(T(A \Rightarrow B) + \varepsilon)
\tag{5}
\end{equation}

In preliminary experiments, this independent logic consistency loss had only a limited impact on robustness, largely because each symbolic attribute was enforced in isolation. This led us to introduce a joint rule loss that treats the full set of symbolic attributes for a class as a single structured constraint, penalizing predictions whose combined attributes do not support the predicted label. Rather than checking shape, color, category, and icon type separately, the joint loss evaluates whether the entire attribute configuration is logically consistent with the class. In this way, it extends the basic logic loss from individual soft implications to a unified multi-attribute consistency constraint. The joint symbolic logic loss is defined as:
\begin{equation}
\begin{split}
\mathcal{L}_{\text{joint}} = -\log(T_{\text{joint}} + \varepsilon)\\
T_{\text{joint}} = \min\left( \frac{\prod_j p_j^{w_j}}{p_{\text{cls}} + \varepsilon},\ 1 \right)
\end{split}
\tag{6}
\end{equation}

Here, \( p_j \) is the predicted probability for the symbolic attribute \( j \), and \( w_j \) is a tunable importance weight. The numerator captures weighted symbolic consistency, and the denominator ensures normalization relative to class confidence. To make logic loss more adaptive, we update the weights $w_j$ in Equation~(6) at each epoch based on the symbolic accuracy of the model. Components with lower accuracy (e.g., shape, text, or icon) receive higher weights, which encourages the model to focus more on correcting those inconsistencies during training.

Furthermore, through these adaptive weights $w_j$, since the loss automatically allocates more pressure to whichever symbolic components are currently weakest, this encourages the backbone to organize its features so that each traffic-sign class occupies a distinct, rule-consistent region in the shared representation space rather than relying on brittle, low-level cues.
Finally, we combine symbolic regularization with standard supervised learning using total loss.
\begin{equation}
\mathcal{L}_{\text{total}} = \mathcal{L}_{\text{CE}} + \lambda_{\text{logic}} \cdot \mathcal{L}_{\text{logic}} + \lambda_{\text{semantic}} \cdot \mathcal{L}_{\text{semantic}}
\tag{7}
\end{equation}
where $\mathcal{L}_{\text{CE}}$ is classification cross-entropy, \( \lambda_{\text{logic}} \) and \( \lambda_{\text{semantic}} \) are hyperparameters controlling the influence of logic and semantic consistency terms. In our formulation, the logic term $L_{\text{logic}}$ corresponds to the joint symbolic loss defined in Equation~(6). We first introduce the basic soft implication loss in Equation~(5) for clarity, and then extend it into the unified joint symbolic loss in Equation~(6), which is the version used during training.
By integrating symbolic rules through soft constraints and semantic regularization, our framework improves both interpretability and adversarial robustness.

\section{Backbone Modification and Symbolic Head Design}
To enable simultaneous prediction of traffic sign classes and their symbolic attributes (such as shape, color, and icon type), we adapt the standard ResNet18 architecture as a shared backbone for both semantic classification and rule-based supervision. Specifically, we make two structural changes to preserve spatial detail in the early feature maps: the initial 7×7 convolution is replaced with a 3×3 kernel, and the max pooling layer is removed. This modification reduces early downsampling and preserves a higher spatial resolution, an essential factor for recognizing fine-grained small traffic signs.
We also remove the final fully connected (FC) classification layer and replace it with an identity mapping. As a result, ResNet produces a 512-dimensional feature vector from its final global average pooling layer. This vector, denoted as \( \mathbf{f} \in \mathbb{R}^{512} \), serves as the input to all downstream heads.
This shared feature vector \( \mathbf{f} \) is passed simultaneously to:
\begin{itemize}
    \item a standard classification head for traffic sign recognition, and
    \item multiple symbolic attribute heads, each implemented as a lightweight multi-layer perceptron (MLP).
\end{itemize}

Each symbolic MLP head consists of:
\begin{itemize}
    \item A linear projection layer that reduces dimensionality (512 → 128),
    \item Batch normalization,
    \item ReLU activation, and
    \item A final linear layer that maps to the output space of a specific symbolic attribute (e.g., number of shapes or color classes).
\end{itemize}
Formally, each symbolic head can be written as:
\[
\text{MLP}_{\text{attr}}(\mathbf{f}) = W_2 \cdot \text{ReLU}(\text{BN}(W_1 \cdot \mathbf{f}))
\]
where \( W_1 \in \mathbb{R}^{128 \times 512} \), \( W_2 \in \mathbb{R}^{d \times 128} \), and \( d \) is the number of output classes for the attribute (e.g., 5 shapes or 6 colors).
We emphasize that no symbolic features are manually engineered or fed as input. All representations are learned directly from the raw images. Symbolic supervision is introduced through auxiliary outputs and corresponding logic-based losses, while the model remains fully end-to-end and differentiable.
Figure~\ref{fig:resnet-arch} illustrates the modified architecture. The shared modified ResNet backbone extracts a global feature vector, which is then used by parallel heads for class and symbolic predictions.

\section{Training Strategy with Symbolic and Adversarial Supervision}
\begin{figure*}[t]
  \centering
  \includegraphics[width=0.8\textwidth]{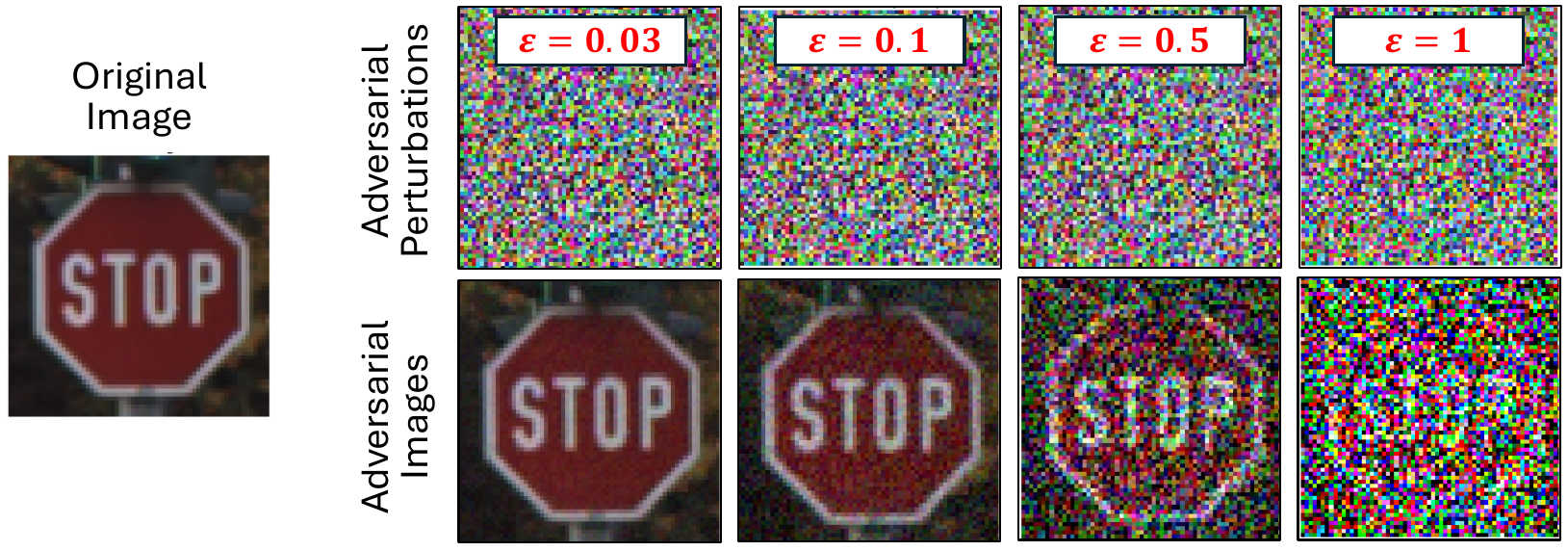}\vspace{-0.5em}
  \caption{Original traffic sign image alongside adversarial perturbations and the corresponding adversarial examples generated at different perturbation budgets ($\varepsilon = 0.03$, $0.1$, $0.5$, and $1$). The perturbation maps illustrate how increasing $\varepsilon$ intensifies pixel-level noise, while the resulting adversarial images show progressively severe corruption of the underlying sign.}
  \label{fig:samples}
\end{figure*}

We train our model on the full German Traffic Sign Recognition Benchmark (GTSRB)~\cite{ref16} dataset using both clean and adversarial examples, guided by a combination of standard classification objectives and symbolic rule supervision. Each input is processed through the modified ResNet18 backbone and encoded into a 512-dimensional feature vector, which is shared between a classifier head and multiple symbolic heads to predict attributes such as shape, color components, category, and icon type.
To increase robustness and interpretability, we jointly train the model using clean and adversarial samples. For each batch, adversarial examples are generated on-the-fly using the FGSM or PGD with a perturbation strength of \( \varepsilon = 8/255 \). This value provides a balanced adversarial threat: large enough to meaningfully perturb the input while remaining visually imperceptible under normalized image scales. Figure~\ref{fig:samples} visualizes how adversarial perturbations of increasing \( \varepsilon\) distort the input image, while remaining largely imperceptible at lower budgets.
The model is supervised using three complementary loss functions: cross-entropy for classification accuracy, semantic loss to enforce symbolic class equivalence, and logic consistency loss to enforce adherence to symbolic rules. These losses are applied to both clean and adversarial samples, each using the formulation defined in Equation~(7).

The joint rule loss plays a particularly important role once adversarial training is introduced. Standard FGSM/PGD training only constrains the class logits to be stable under perturbations, which still allows the backbone to route gradients through brittle, non-semantic features as long as the top-1 label remains correct. In our framework, every adversarial example is simultaneously evaluated by the classifier head and the symbolic MLP heads. An adversarial input that preserves the predicted class label but causes one or more symbolic attributes (e.g., shape or category) to take inconsistent values incurs a large joint logic loss. Because all symbolic heads share the same 512-dimensional backbone representation, the gradients from $L_{\text{joint}}$ force the backbone to organize its features so that robustness is achieved through semantically consistent attribute configurations, not in spite of them. In other words, the joint rule loss does not by itself create robustness, but when combined with adversarial training it disciplines how robustness is expressed, steering the network toward decision regions where both class predictions and symbolic attributes remain jointly stable under attack.

\begin{figure}[t]
  \centering
  \includegraphics[width=0.8\linewidth]{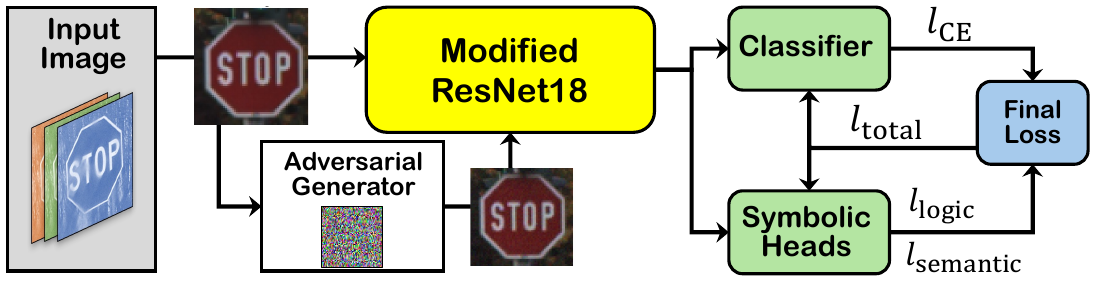}
  \caption{Illustration of our training strategy. Each input image is used to generate an adversarial example, and both clean and adversarial samples are passed through a shared modified ResNet18 backbone followed by the classifier and symbolic heads. The final loss combines classification cross-entropy with semantic and symbolic-logic losses, which jointly update the model.}
  \label{fig:training_pipeline}
\end{figure}

This joint training scheme ensures that the model learns to classify correctly under both natural and adversarial conditions, while also preserving symbolic coherence in its predictions. Unlike conventional adversarial training that focuses purely on classification robustness, our approach additionally penalizes violations of symbolic logic, yielding improved structure-aware generalization. Figure~\ref{fig:training_pipeline} illustrates our training pipeline. Each input image is first passed to the adversarial generator to produce its adversarial counterpart. Both the clean and adversarial samples are then fed into the modified ResNet18 backbone, followed by lightweight MLP heads, which output the semantic and symbolic predictions. The model is supervised on clean and adversarial samples using classification, semantic, and symbolic logic losses, and the total loss is derived from the weighted sum of these components. This architecture ensures robustness and logical consistency under both standard and adversarial conditions.
To focus on symbolic regularization where it is most needed, we compute an adaptive logic weight \( \lambda_{\text{logic}} \) for each batch, based on classification cross-entropy and prediction correctness. This increases logic pressure on uncertain or misclassified samples, while reducing it when the model is confident and accurate. In contrast, the semantic loss weight \( \lambda_{\text{semantic}} \) remains fixed, as its regularization effect is globally beneficial regardless of the certainty of the prediction. 
Finally, while symbolic reasoning introduces additional computational cost compared to standard adversarial training, the resulting improvements in interpretability and adversarial robustness justify the overhead, especially in safety-critical applications like traffic sign recognition.

\section{Evaluation}
The proposed \DesignI is evaluated on the GTSRB dataset. The model is trained from scratch for only 10 epochs, without any pre-training. This deliberate limitation serves two purposes. First, GTSRB has highly structured visuals, which makes models converge quickly; training for too long can therefore lead to overfitting to low-level features. Second, our goal is to encourage the model to rely more on symbolic reasoning and structural supervision, rather than memorizing dataset-specific patterns. Despite this short training schedule, our model achieves strong generalization, demonstrating the effectiveness of symbolic regularization.
We evaluated the performance of our model in terms of both adversarial robustness and inference time computation overhead. First, we compare it with standard adversarial training methods, showing that the integration of symbolic reasoning significantly improves robustness under FGSM and PGD attacks while maintaining clean accuracy. We also benchmark our approach against a recent transformer-based defense method, demonstrating competitive or superior accuracy with much lower training cost and better interpretability. Finally, we analyze the inference-time computation overhead of our neuro-symbolic model, highlighting a favorable trade-off between robustness and efficiency, and discuss strategies for further reducing computational complexity in resource-constrained environments.
\begin{table*}[t]
\centering
\caption{Comparison of clean and adversarial accuracy across different training strategies.}
\begin{tabular}{lccc}
\hline
\textbf{Training Method} & \textbf{Clean Accuracy}* & \textbf{FGSM Accuracy}* & \textbf{PGD Accuracy}* \\
\hline \hline
No Adversarial Training & ~98.67\% & ~36.7\% & ~25.5\% \\
FGSM Training Only & ~96.5\% & ~45.0\% & ~35.0\% \\
PGD Training Only & ~4.5\% & ~43.6\% & ~38.65\% \\
FGSM-Clean Training & ~96.40\% & ~46.70\% & ~34.10\% \\
PGD-Clean Training & ~7.00\% & ~44.3\% & ~39.70\% \\
Neuro-Symbolic (ours) & ~97.90\% & ~38.30\% & ~19.90\% \\
FGSM-Neuro-Symboli`c (ours) & ~98.20\% & ~63.10\% & ~42.68\% \\
PGD-Neuro-Symbolic (ours) & ~98.5\% & ~65.75\% & ~56.00\% \\
\hline
\end{tabular}
 
\begin{flushleft}
\centering
*\scriptsize{\textbf{Clean Accuracy} = no attack; \textbf{FGSM Accuracy} = accuracy under the FGSM attack; \textbf{PGD Accuracy} = accuracy under the PGD attack}
\end{flushleft}
\label{tab:adv_results}
\end{table*}

\subsection{Comparison Results}
\subsubsection{Ours versus Standard Adversarial Training}
We evaluated the robustness of the model under three conditions: standard clean test inputs, adversarial examples generated using FGSM, and those generated by the PGD attack. For both attack methods, we use a perturbation strength of \( \varepsilon = 8/255 \), which controls the maximum allowable change to each input pixel. This value is chosen to meaningfully distort the input while remaining visually imperceptible, thereby posing a realistic adversarial threat.
Table~\ref{tab:adv_results} summarizes the results in several training configurations.
The first column of this table reports test accuracy on clean samples, while the second and third columns show robustness under FGSM and PGD attacks, respectively. Our neuro-symbolic model combined with adversarial training (last two rows) achieves the highest accuracy under adversarial examples, while still maintaining over 98\% accuracy on clean (non-adversarial) inputs, comparable to the best nonadversarial baseline.

For example, compared to FGSM adversarial training (45.0\% FGSM accuracy), FGSM-Neuro-Symbolic achieves 63.10\%, an improvement of +18.1\%. Similarly, PGD-Neuro-Symbolic improves PGD accuracy from 38.65\% (PGD adversarial training) to 56.00\%, a gain of +17.35\%. This shows that symbolic rule supervision significantly enhances robustness beyond what standard adversarial training achieves.
Our pure Neuro-Symbolic model (without any adversarial training) maintains high clean accuracy ($\sim97.90\%$) but drops under attack, especially PGD ($\sim19.90\%$). This validates that symbolic logic alone is not enough. However, when combined with adversarial objectives, the hybrid model outperforms all other strategies under attack, demonstrating the complementarity of logical reasoning and adversarial supervision.

\begin{figure}[h]
  \centering
  \includegraphics[width=0.6\linewidth]{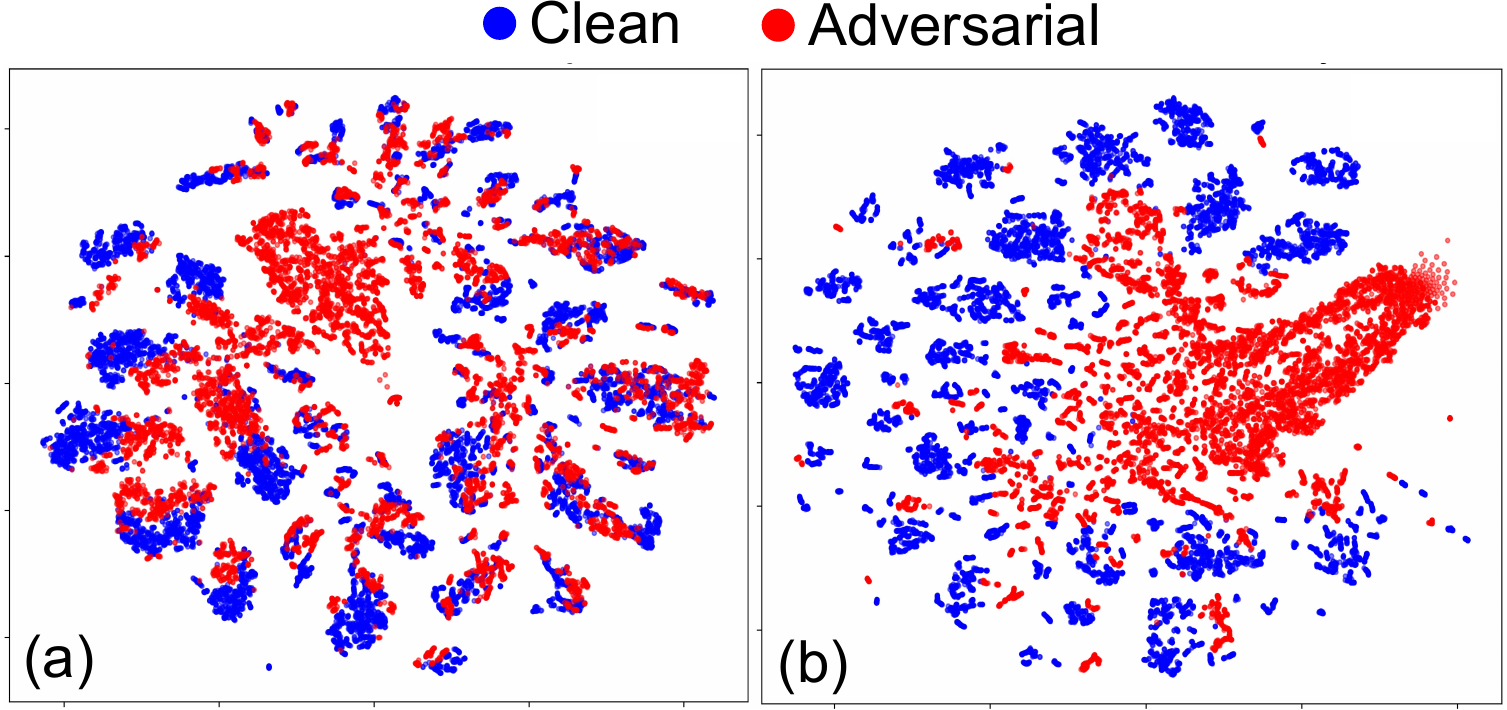}
  \caption{Comparison of clean and adversarial sample distributions using t-SNE projections. (a) Our Neuro-Symbolic model better preserves semantic structure, with adversarial samples remaining closely aligned with their clean counterparts, highlighting the benefits of rule-based reasoning in maintaining robustness. (b) Baseline ResNet model shows disrupted structure and significant overlap between clean and adversarial samples.}
  \label{fig:tsne_comparison}
\end{figure}

Interestingly, PGD-Clean  achieves a very low clean accuracy ($\sim7.00\%$), likely due to the inclusion of strong adversarial signals that distort the clean data representation. Although it maintains reasonable PGD accuracy ($\sim39.70\%$), it underperforms compared to our hybrid model.
Figure~\ref{fig:tsne_comparison}(a) shows the distribution of clean and adversarial samples for our hybrid Neuro-Symbolic model. Adversarial samples remain more closely grouped with their clean counterparts, indicating that the reasoning module helps preserve semantic structure and supports robust, rule-consistent predictions even under attack. In contrast, Figure~\ref{fig:tsne_comparison}(b) visualizes the distribution for a baseline ResNet model. The adversarial perturbations disrupt the structure of the internal feature space, leading to poor clustering and significant overlap between clean and adversarial samples.

\subsubsection{Ours versus Transformer-Based Models}
We further compare our neuro-symbolic framework with \textit{LNL-MoEx}~\cite{ref17}, a recent transformer-based defense method that integrates locality inductive bias with moment exchange (MoEx) to enhance adversarial robustness. LNL-MoEx is trained on GTSRB for 100 epochs and on CIFAR-10 for 150 epochs using a ViT-based architecture, and leverages MoEx as a form of statistical feature augmentation.
Importantly, while many ViT-based defenses rely on ImageNet pre-training, the authors of LNL-MoEx emphasize that their method is trained from scratch on CIFAR-10 and GTSRB without ImageNet pre-training.

Table~\ref{tab:adv_comparison_transformer} presents a comparative analysis of adversarial accuracy under FGSM and PGD attacks. With \( \varepsilon = 8/255~\sim~0.03 \), our PGD-Neuro-Symbolic model achieves 56.00\% PGD accuracy, slightly outperforming LNL-MoEx-Ti (55.4\%) and approaching the best-performing transformer baseline, LNL-MoEx-S (59.4\%), despite requiring no data augmentation and only 10 training epochs. Under FGSM attack with \( \varepsilon = 8/255 \), our model reaches 63.10\% accuracy, which is competitive with LNL-S (64.5\%) and significantly better than LNL-Ti (57.7\%). Although LNL-MoEx-S achieves the highest FGSM accuracy (77.8\%), it relies on transformer backbones and extensive data augmentation~\cite{ref17}. But the accuracy under PGD attack is more meaningful, because PGD performs multiple iterative gradient steps within the same $\varepsilon$-constraint, producing more effective and adversarially optimal perturbations than the single-step FGSM attack. Consequently, PGD robustness is widely regarded as the more challenging and informative benchmark.

It is important to note that LNL-MoEx evaluates robustness using perturbation strengths up to $\varepsilon=1$, a very large budget that does not reflect realistic threat levels for safety-critical tasks such as traffic sign recognition. To enable a fairer comparison within their perturbation scale, we trained an FGSM-Neuro-Symbolic variant at $\varepsilon=0.5$, which achieves 74.00\% FGSM accuracy, closing the gap with LNL-MoEx-S (77.8\%) despite our simpler architecture and minimal training overhead. However, when training a PGD-based Neuro-Symbolic variant at $\varepsilon=0.5$, the optimization became unstable and failed to converge, indicating that very large-magnitude perturbations may require additional stabilization techniques. Although such a large perturbation budget is uncommon in standard threat models—and, as shown in Figure~\ref{fig:samples}, large values of $\varepsilon$ heavily corrupt the input image—we use $\varepsilon = 0.5$ solely for fair comparison with LNL-MoEx.

\begin{table}[t]
\centering
\caption{Adversarial accuracy under FGSM and PGD attacks for different methods. Our neuro-symbolic variants are compared with transformer-based LNL~\cite{ref17} models with and without MoEx data augmentation.}
\scalebox{0.92}{
\begin{tabular}{lcc}
\hline
\textbf{Method} & \textbf{FGSM Accuracy} & \textbf{PGD Accuracy} \\
\hline \hline
LNL-Ti~\cite{ref17} & 57.7\% & 37.9\% \\
LNL-S~\cite{ref17} & 64.5\% & 45.7\% \\
\begin{tabular}[c]{@{}l@{}}LNL-MoEx-Ti~\cite{ref17}\\ (Data Augmentation)\end{tabular} & 71.3\% & 55.4\% \\
\begin{tabular}[c]{@{}l@{}}LNL-MoEx-S~\cite{ref17}\\ (Data Augmentation)\end{tabular} & 77.8\% & 59.4\% \\
FGSM-Neuro-Symbolic ($\varepsilon=0.03$) & 63.10\% & 42.68\% \\
FGSM-Neuro-Symbolic ($\varepsilon=0.5$) & 74.00\% & 25.00\% \\
PGD-Neuro-Symbolic ($\varepsilon=0.03$) & 65.75\% & 56.00\% \\
\hline
\end{tabular}
}
\label{tab:adv_comparison_transformer}
\end{table}

Compared to LNL-MoEx, our model offers several advantages:
\begin{itemize}
    \item \textbf{Interpretability}: While LNL-MoEx enhances robustness through architectural and statistical techniques, our neuro-symbolic model improves both robustness and interpretability by enforcing semantic rule consistency that provides transparent reasoning behind each prediction.
    \item \textbf{Lightweight Regularization}: Our symbolic supervision module is orthogonal to the model architecture and can be easily added to any neural backbone without architectural redesign or tuning of fusion strategies.
    \item \textbf{Efficient Training}: Our model achieves competitive adversarial performance using only 10 training epochs on GTSRB, while LNL-MoEx requires over 100 epochs and complex MoEx augmentation to reach similar results.
\end{itemize}
To further contextualize our results, Table~\ref{tab:transformer_comparison} compares the adversarial performance of our neuro-symbolic variants against a wide range of transformer-based models with and without data augmentation, as reported in previous work~\cite{ref17}. In particular, many of these transformer baselines exhibit low robustness under FGSM and PGD attacks when trained without augmentations. Even with advanced augmentation strategies such as Puzzle-Mix and DeepAugment, our PGD-Neuro-Symbolic model achieves superior or comparable robustness (56.00\% under PGD), while maintaining a lightweight architecture and minimal training cost. This highlights the efficacy of symbolic reasoning in enhancing robustness, without relying on large-scale transformer backbones or augmentation pipelines.

\begin{table}[t]
\centering
\caption{Comparison of adversarial accuracy under FGSM and PGD attacks across various transformer-based models with and without data augmentations, alongside our neuro-symbolic variants.}
\scalebox{0.92}{
\begin{tabular}{lcc}
\hline
\textbf{Method} & \textbf{FGSM Accuracy} & \textbf{PGD Accuracy} \\
\hline \hline
\multicolumn{3}{c}{\textsl{Without Augmentations}} \\
\hline
PVT-Tiny~\cite{ref18} & 10.8\% & 2.0\% \\
TNT-T~\cite{ref19} & 10.9\% & 3.9\% \\
T2T-ViT-t-10~\cite{ref20} & 10.2\% & 0.4\% \\
RVT-Ti~\cite{ref21} & 31.5\% & 13.9\% \\
Swin-T~\cite{ref22} & 20.2\% & 7.8\% \\
PVT-Small~\cite{ref18} & 21.4\% & 2.4\% \\
TNT-S~\cite{ref19} & 21.2\% & 6.8\% \\
T2T-ViT-t-14~\cite{ref20} & 21.7\% & 4.9\% \\
RVT-S~\cite{ref21} & 46.1\% & 25.3\% \\
\hline
\multicolumn{3}{c}{\textsl{With Augmentations}} \\
\hline
DeepAugment~\cite{ref23} & \textbf{65.1\%} & 42.2\% \\
CutMix~\cite{ref24} & 62.6\% & 40.1\% \\
AugMix~\cite{ref25} & 61.7\% & 39.8\% \\
Puzzle-Mix~\cite{ref26} & \textbf{67.9\%} & \textbf{44.2\%} \\
RVT-Ti*~\cite{ref21} & 38.9\% & 20.7\% \\
RVT-S*~\cite{ref21} & 51.2\% & 32.9\% \\
\hline
\multicolumn{3}{c}{\textbf{\textsl{Neuro-Symbolic (Ours)}}} \\
\hline
FGSM-Neuro-Symbolic ($\varepsilon=0.03$) & 63.10\% & 42.68\% \\
FGSM-Neuro-Symbolic ($\varepsilon=0.5$) & 74.00\% & 25.00\% \\
PGD-Neuro-Symbolic ($\varepsilon=0.03$) & 65.75\% & 56.00\% \\
\hline
\end{tabular}
}
\label{tab:transformer_comparison}
\end{table}

\subsection{Inference-Time Computation Comparison}
\label{sec:inference_comparison}

While previous sections focused on robustness and accuracy, we now compare the inference-time computational cost of our neuro-symbolic model against both adversarial training and transformer-based LNL-MoEx~\cite{ref17}. As mentioned before, our model uses a modified ResNet18 backbone followed by symbolic heads. This setup enables reasoning about both class predictions and symbolic consistency under perturbation and clean samples. Each symbolic head is a lightweight MLP, as previously described.
The resulting inference-time computation cost is close to that of a standard ResNet18 with additional symbolic heads—much lower than transformer-based defenses.
In contrast, adversarially trained models (e.g., PGD-trained ResNet18) require only a single forward pass and no symbolic reasoning, yielding slightly lower cost but no interpretability. Meanwhile, LNL-MoEx uses a ViT-based backbone and incorporates moment exchange logic and locality-aware constraints, leading to substantially higher inference-time complexity.

\begin{table}[t]
\centering
\caption{Comparison of inference-time computational cost across architectures. Our model balances reasoning and robustness with moderate overhead.}
\scalebox{0.85}{
\begin{tabular}{lccc}
\hline \hline
\textbf{Method} & \textbf{Backbone} & \textbf{Accuracy(PGD)} & \textbf{Inference Cost} \\
\hline
Adversarial Training & ResNet18 & 38.65\% & Very Low \\
LNL-MoEx~\cite{ref17} & ViT & 59.4\% & High \\
\DesignII & ResNet18 + MLPs & 56.0\% & Low \\
\hline
\end{tabular}
}
\label{tab:inference_comparison}
\end{table}

Despite relying on symbolic reasoning, our model remains significantly more efficient than transformer-based defenses. As shown in Table~\ref{tab:inference_comparison}, our approach achieves a practical trade-off—maintaining strong robustness and logic-driven interpretability, with only a low increase in computation relative to standard ResNet18.
As traffic sign recognition systems are integral components of autonomous driving pipelines, their deployment on devices with limited resources and constrained computational budgets is a critical consideration. To this end, reducing inference-time computation overhead is essential for practical, real-world adoption. One promising direction is to explore techniques such as \textit{early exit mechanisms} or \textit{model quantization}, as suggested in prior work on neural network optimization for edge deployment~\cite{ref27,solanki2025atm}, to reduce computational complexity further while maintaining symbolic robustness and interpretability.

\section{Conclusion}
This paper introduced \DesignII, a neuro-symbolic framework that enhances the adversarial robustness and interpretability of deep learning models by integrating symbolic rule supervision. Through extensive experiments on the GTSRB dataset, we demonstrated that our model outperforms both standard adversarial training and state-of-the-art transformer-based defense methods, while requiring fewer training epochs and no data augmentation. Our PGD-Neuro-Symbolic model achieved acceptable accuracy under FGSM and PGD attacks while maintaining its accuracy on the clean dataset. Additionally, our approach balances robustness and reasoning with moderate inference-time computation overhead, making it a promising candidate for real-world deployment.
In future work, we aim to further adapt our framework for edge devices and energy-constrained environments, such as those used in autonomous vehicles, by exploring lightweight backbones, early exits, and quantized inference. Another direction is to extend symbolic reasoning beyond static rule checks toward chained logical inference, enabling the model to apply domain knowledge to dynamic driving scenarios and generate new symbolic rules from foundational ones. This would allow neuro-symbolic systems to assist in higher-level decision-making under uncertainty, moving toward generalizable and trustworthy AI in safety-critical applications.

\section*{Data Availability}
The experiments in this study use the GTSRB dataset, which is publicly available through the following \href{https://www.kaggle.com/datasets/meowmeowmeowmeowmeow/gtsrb-german-traffic-sign?}{Kaggle repository}. This dataset provides the raw training and testing images used in our experiments. All adversarial data (FGSM and PGD) used in this work were generated on the fly by the algorithms implemented in the accompanying code repository, and no additional datasets were created or hosted externally. The full code for generating this data and reproducing our experiments is available on the  \href{https://github.com/iDEALabAcademy/NeuroShield}{GitHub repository}. Additional experimental artifacts (e.g., trained model checkpoints and logs) are available from the corresponding author upon reasonable request.

\bibliography{ref}

@inproceedings{ref28,
  title={SHIELDeNN: Online accelerated framework for fault-tolerant deep neural network architectures},
  author={Khoshavi, Navid and Roohi, Arman and Broyles, Connor and Sargolzaei, Saman and Bi, Yu and Pan, David Z},
  booktitle={2020 57th ACM/IEEE Design Automation Conference (DAC)},
  pages={1--6},
  year={2020},
  organization={IEEE}
}

@article{ref1,
  title={Efficient processing of deep neural networks: A tutorial and survey},
  author={Sze, Vivienne and Chen, Yu-Hsin and Yang, Tien-Ju and Emer, Joel S},
  journal={Proceedings of the IEEE},
  volume={105},
  number={12},
  pages={2295--2329},
  year={2017},
  publisher={Ieee}
}

@article{ref2,
    title={Logic tensor networks},
  author={Badreddine, Samy and Garcez, Artur d'Avila and Serafini, Luciano and Spranger, Michael},
  journal={Artificial Intelligence},
  volume={303},
  pages={103649},
  year={2022},
  publisher={Elsevier}
}

@article{ref3,
  title={A survey on adversarial attacks and defences},
  author={Chakraborty, Anirban and Alam, Manaar and Dey, Vishal and Chattopadhyay, Anupam and Mukhopadhyay, Debdeep},
  journal={CAAI Transactions on Intelligence Technology},
  volume={6},
  number={1},
  pages={25--45},
  year={2021},
  publisher={Wiley Online Library}
}

@article{ref5,
  title={Towards data-and knowledge-driven AI: a survey on neuro-symbolic computing},
  author={Wang, Wenguan and Yang, Yi and Wu, Fei},
  journal={IEEE Transactions on Pattern Analysis and Machine Intelligence},
  year={2024},
  publisher={IEEE}
}

@article{ref6,
  title={Neural logic machines},
  author={Dong, Honghua and Mao, Jiayuan and Lin, Tian and Wang, Chong and Li, Lihong and Zhou, Denny},
  journal={arXiv preprint arXiv:1904.11694},
  year={2019}
}

@article{ref7,
title={Neuro-symbolic forward reasoning},
  author={Shindo, Hikaru and Dhami, Devendra Singh and Kersting, Kristian},
  journal={arXiv preprint arXiv:2110.09383},
  year={2021}
}

@inproceedings{ref8,
  title={A semantic loss function for deep learning with symbolic knowledge},
  author={Xu, Jingyi and Zhang, Zilu and Friedman, Tal and Liang, Yitao and Broeck, Guy},
  booktitle={International conference on machine learning},
  pages={5502--5511},
  year={2018},
  organization={PMLR}
}

@article{ref9,
  title={Neuro-symbolic verification of deep neural networks},
  author={Xie, Xuan and Kersting, Kristian and Neider, Daniel},
  journal={arXiv preprint arXiv:2203.00938},
  year={2022}
}

@article{ref10,
  title={Explaining and harnessing adversarial examples},
  author={Goodfellow, Ian J and Shlens, Jonathon and Szegedy, Christian},
  journal={arXiv preprint arXiv:1412.6572},
  year={2014}
}

@article{ref11,
  title={Towards deep learning models resistant to adversarial attacks},
  author={Madry, Aleksander and Makelov, Aleksandar and Schmidt, Ludwig and Tsipras, Dimitris and Vladu, Adrian},
  journal={arXiv preprint arXiv:1706.06083},
  year={2017}
}

@inproceedings{ref12,
  title={Obfuscated gradients give a false sense of security: Circumventing defenses to adversarial examples},
  author={Athalye, Anish and Carlini, Nicholas and Wagner, David},
  booktitle={International conference on machine learning},
  pages={274--283},
  year={2018},
  organization={PMLR}
}

@inproceedings{ref13,
  title={Adversarial examples are not easily detected: Bypassing ten detection methods},
  author={Carlini, Nicholas and Wagner, David},
  booktitle={Proceedings of the 10th ACM workshop on artificial intelligence and security},
  pages={3--14},
  year={2017}
}

@inproceedings{ref14,
  title={Certified adversarial robustness via randomized smoothing},
  author={Cohen, Jeremy and Rosenfeld, Elan and Kolter, Zico},
  booktitle={international conference on machine learning},
  pages={1310--1320},
  year={2019},
  organization={PMLR}
}

@inproceedings{ref15,
  title={Pylon: A pytorch framework for learning with constraints},
  author={Ahmed, Kareem and Li, Tao and Ton, Thy and Guo, Quan and Chang, Kai-Wei and Kordjamshidi, Parisa and Srikumar, Vivek and Van den Broeck, Guy and Singh, Sameer},
  booktitle={NeurIPS 2021 Competitions and Demonstrations Track},
  pages={319--324},
  year={2022},
  organization={PMLR}
}

@inproceedings{ref16,
  title={The German traffic sign recognition benchmark: a multi-class classification competition},
  author={Stallkamp, Johannes and Schlipsing, Marc and Salmen, Jan and Igel, Christian},
  booktitle={The 2011 international joint conference on neural networks},
  pages={1453--1460},
  year={2011},
  organization={IEEE}
}

@article{ref17,
  title={Robust transformer with locality inductive bias and feature normalization},
  author={Manzari, Omid Nejati and Kashiani, Hossein and Dehkordi, Hojat Asgarian and Shokouhi, Shahriar B},
  journal={Engineering Science and Technology, an International Journal},
  volume={38},
  pages={101320},
  year={2023},
  publisher={Elsevier}
}

@inproceedings{ref18,
  title={Pyramid vision transformer: A versatile backbone for dense prediction without convolutions},
  author={Wang, Wenhai and Xie, Enze and Li, Xiang and Fan, Deng-Ping and Song, Kaitao and Liang, Ding and Lu, Tong and Luo, Ping and Shao, Ling},
  booktitle={Proceedings of the IEEE/CVF international conference on computer vision},
  pages={568--578},
  year={2021}
}

@article{ref19,
  title={Transformer in transformer},
  author={Han, Kai and Xiao, An and Wu, Enhua and Guo, Jianyuan and Xu, Chunjing and Wang, Yunhe},
  journal={Advances in neural information processing systems},
  volume={34},
  pages={15908--15919},
  year={2021}
}

@inproceedings{ref20,
  title={Tokens-to-token vit: Training vision transformers from scratch on imagenet},
  author={Yuan, Li and Chen, Yunpeng and Wang, Tao and Yu, Weihao and Shi, Yujun and Jiang, Zi-Hang and Tay, Francis EH and Feng, Jiashi and Yan, Shuicheng},
  booktitle={Proceedings of the IEEE/CVF international conference on computer vision},
  pages={558--567},
  year={2021}
}

@inproceedings{ref21,
  title={Towards robust vision transformer},
  author={Mao, Xiaofeng and Qi, Gege and Chen, Yuefeng and Li, Xiaodan and Duan, Ranjie and Ye, Shaokai and He, Yuan and Xue, Hui},
  booktitle={Proceedings of the IEEE/CVF conference on Computer Vision and Pattern Recognition},
  pages={12042--12051},
  year={2022}
}

@inproceedings{ref22,
  title={Swin transformer: Hierarchical vision transformer using shifted windows},
  author={Liu, Ze and Lin, Yutong and Cao, Yue and Hu, Han and Wei, Yixuan and Zhang, Zheng and Lin, Stephen and Guo, Baining},
  booktitle={Proceedings of the IEEE/CVF international conference on computer vision},
  pages={10012--10022},
  year={2021}
}

@inproceedings{ref23,
  title={The many faces of robustness: A critical analysis of out-of-distribution generalization},
  author={Hendrycks, Dan and Basart, Steven and Mu, Norman and Kadavath, Saurav and Wang, Frank and Dorundo, Evan and Desai, Rahul and Zhu, Tyler and Parajuli, Samyak and Guo, Mike and others},
  booktitle={Proceedings of the IEEE/CVF international conference on computer vision},
  pages={8340--8349},
  year={2021}
}

@inproceedings{ref24,
  title={Cutmix: Regularization strategy to train strong classifiers with localizable features},
  author={Yun, Sangdoo and Han, Dongyoon and Oh, Seong Joon and Chun, Sanghyuk and Choe, Junsuk and Yoo, Youngjoon},
  booktitle={Proceedings of the IEEE/CVF international conference on computer vision},
  pages={6023--6032},
  year={2019}
}

@article{ref25,
  title={Augmix: A simple data processing method to improve robustness and uncertainty},
  author={Hendrycks, Dan and Mu, Norman and Cubuk, Ekin D and Zoph, Barret and Gilmer, Justin and Lakshminarayanan, Balaji},
  journal={arXiv preprint arXiv:1912.02781},
  year={2019}
}

@inproceedings{ref26,
  title={Puzzle mix: Exploiting saliency and local statistics for optimal mixup},
  author={Kim, Jang-Hyun and Choo, Wonho and Song, Hyun Oh},
  booktitle={International conference on machine learning},
  pages={5275--5285},
  year={2020},
  organization={PMLR}
}

@inproceedings{ref27,
  title={RL-SEP: RL-Based S mart E xit Point Selection for Enhancing Energy Harvested System Longevity},
  author={Shafiee Sarvestani, Ali and Tabrizchi, Sepehr and Sehatbakhsh, Nader and Roohi, Arman},
  booktitle={Proceedings of the 23rd ACM Conference on Embedded Networked Sensor Systems},
  pages={638--639},
  year={2025}
}

@article{solanki2025atm,
  title={ATM-Net: Adaptive Termination and Multi-Precision Neural Networks for Energy-Harvested Edge Intelligence},
  author={Solanki, Neeraj and Tabrizchi, Sepehr and Sohrabi, Samin and Schmidt, Jason and Roohi, Arman},
  journal={arXiv preprint arXiv:2502.09822},
  year={2025}
}

@article{taha2002symbolic,
  title     = {Symbolic Interpretation of Artificial Neural Networks},
  author    = {Taha, Ismail A. and Ghosh, Joydeep},
  journal   = {IEEE Transactions on Knowledge and Data Engineering},
  volume    = {11},
  number    = {3},
  pages     = {448--463},
  year      = {1999},
  publisher = {IEEE}
}



\section*{Funding}
This work is supported in part by the National Science Foundation (NSF) under grant numbers 2504839, 2447566, and 2448133.

\section*{Author contributions statement}
A. Sh. conceived the main idea and led the initial study design. A. Sh. and J. S. implemented the simulations, carried out the analyses, and prepared the initial manuscript draft. J. S. further refined and interpreted the results. A. R. contributed to developing the concept, critically reviewed the simulations and analyses, discussed and interpreted the findings, revised and proofread the manuscript, and supervised the overall study.




\section*{Declarations}
\section*{Competing interests}
The authors declare no competing interests.
\section*{Additional information}
\textbf{Correspondence} and requests for materials should be addressed to Arman Roohi.

\end{document}